\documentclass[conference]{IEEEtran}
\IEEEoverridecommandlockouts

\usepackage{cite}
\usepackage{amsmath,amssymb,amsfonts}
\usepackage{algorithmic}
\usepackage{graphicx}
\usepackage{textcomp}
\usepackage{xcolor}
\def\BibTeX{{\rm B\kern-.05em{\sc i\kern-.025em b}\kern-.08em
    T\kern-.1667em\lower.7ex\hbox{E}\kern-.125emX}}
    
\usepackage{bibentry}
\usepackage{booktabs}
\usepackage{array}
\usepackage{multirow}
\usepackage{hyperref} 
\usepackage{fontawesome}

\begin{document}

\newcommand{\etc}{\textit{etc}.}
\newcommand{\ie}{\textit{i}.\textit{e}.}
\newcommand{\eg}{\textit{e}.\textit{g}.}

\title{Enhancing Vision-Language Tracking by Effectively Converting Textual Cues into Visual Cues\\
\thanks{
Code: \href{https://github.com/XiaokunFeng/CTVLT}{https://github.com/XiaokunFeng/CTVLT}.

This work is jointly supported by the National Science and Technology Major Project (No.2022ZD0116403), the National Natural Science Foundation of China (No.62176255), and the Strategic Priority Research Program of Chinese Academy of Sciences (No.XDA27010201).}
}

\DeclareRobustCommand*{\IEEEauthorrefmark}[1]{%
  \raisebox{0pt}[0pt][0pt]{\textsuperscript{\footnotesize #1}}%
}
\author{
    \IEEEauthorblockN{
        Xiaokun Feng\IEEEauthorrefmark{1,2}, 
        Dailing Zhang\IEEEauthorrefmark{1,2}, 
        Shiyu Hu\IEEEauthorrefmark{5}, 
        Xuchen Li\IEEEauthorrefmark{1,2}, \\
        Meiqi Wu\IEEEauthorrefmark{3}, 
        Jing Zhang\IEEEauthorrefmark{2}, 
        Xiaotang Chen\IEEEauthorrefmark{1,2,4} and
        Kaiqi Huang\IEEEauthorrefmark{1,2,4}\textsuperscript{(\faEnvelopeO)}
    }
    \IEEEauthorblockA{
        \IEEEauthorrefmark{1}School of Artificial Intelligence, University of Chinese Academy of Sciences, Beijing, China\\
        \IEEEauthorrefmark{2}Institute of Automation, Chinese Academy of Sciences, Beijing, China\\
        \IEEEauthorrefmark{3}School of Computer Science and Technology, University of Chinese Academy of Sciences, Beijing, China\\
        \IEEEauthorrefmark{4}Center for Excellence in Brain Science and Intelligence Technology, Chinese Academy of Sciences, Shanghai, China\\
        \IEEEauthorrefmark{5}School of Physical and Mathematical Sciences, Nanyang Technological University, Singapore\\
        Email: 
            \{fengxiaokun2022, zhangdailing2023\}@ia.ac.cn, shiyu.hu@ntu.edu.sg, lixuchen2024@ia.ac.cn,\\ wumeiqi18@mails.ucas.ac.cn, jing\_zhang@ia.ac.cn, \{xtchen,kaiqi.huang\}@nlpr.ia.ac.cn
    }
}

\maketitle

\begin{abstract}
Vision-Language Tracking (VLT) aims to localize a target in video sequences using a visual template and language description. While textual cues enhance tracking potential, current datasets typically contain much more image data than text, limiting the ability of VLT methods to align the two modalities effectively. To address this imbalance, we propose a novel plug-and-play method named CTVLT that leverages the strong text-image alignment capabilities of foundation grounding models. CTVLT converts textual cues into interpretable visual heatmaps, which are easier for trackers to process. Specifically, we design a textual cue mapping module that transforms textual cues into target distribution heatmaps, visually representing the location described by the text. Additionally, the heatmap guidance module fuses these heatmaps with the search image to guide tracking more effectively. Extensive experiments on mainstream benchmarks demonstrate the effectiveness of our approach, achieving state-of-the-art performance and validating the utility of our method for enhanced VLT.
\end{abstract}


\section{Introduction}

As an extension of the classical Visual Object Tracking (VOT) task \cite{OTB2013} \cite{SOTVerse} \cite{hu2022global} \cite{biodrone}, the Vision-Language Tracking (VLT) task \cite{OTB-Lang} focuses on locating a user-specified object in a video sequence using multimodal inputs, including a visual template and a textual description. By leveraging the complementary strengths of both modalities, Vision-Language Trackers (VLTs) hold the potential for enhanced tracking performance, garnering significant recent attention \cite{hu2023multi}\cite{li2024dtllm}\cite{li2024texts}\cite{feng2024memvlt}. However, incorporating textual cues introduces challenges, particularly in ensuring effective fine-grained alignment between text and
images.

\begin{figure}[htbp]
\centering
\includegraphics[width=\linewidth]{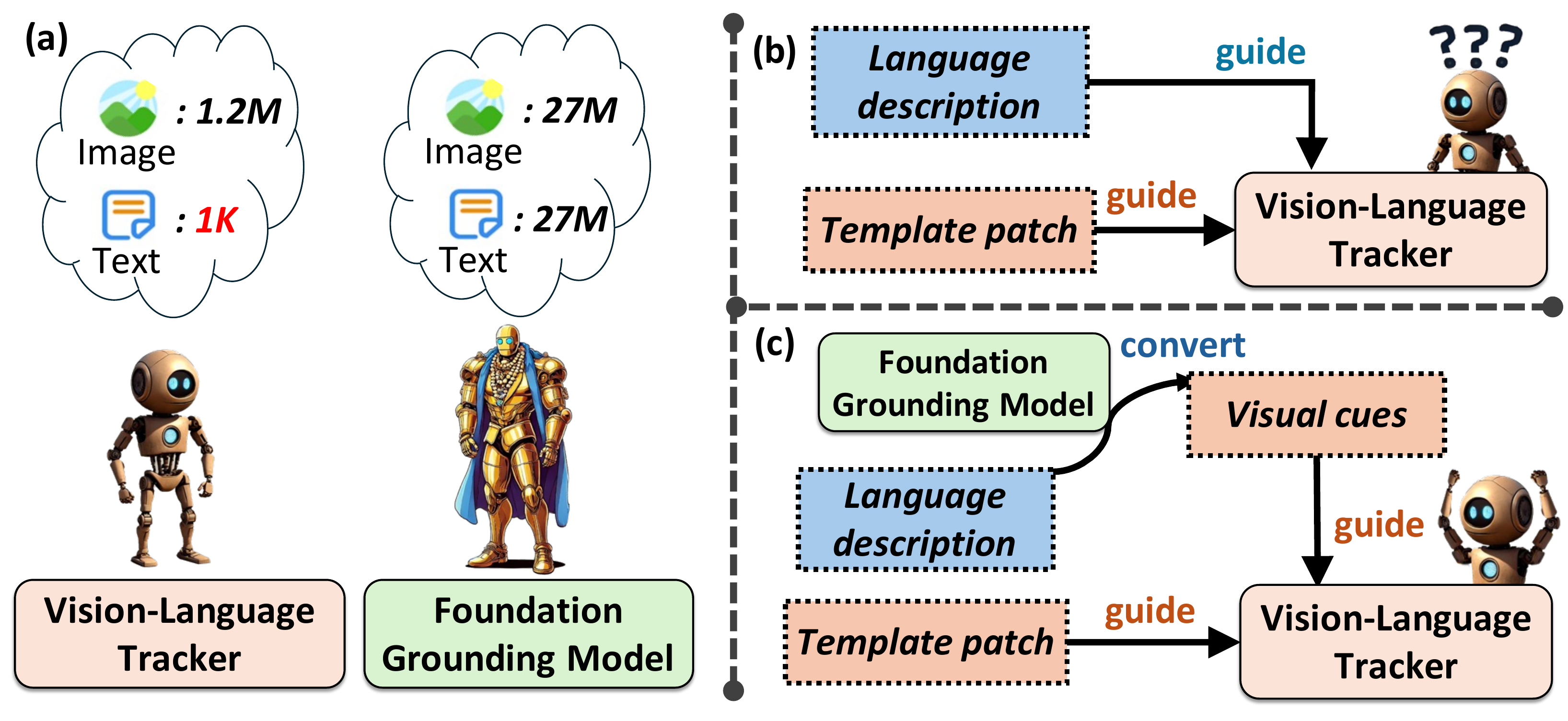}
\caption{
\textbf{Schematic diagram of motivation and method paradigm innovation.}
\textbf{(a):} Comparison of training environments between vision-language trackers and foundation grounding models. 
\textbf{(b):} The severe scarcity of textual data limits the tracker’s ability to understand text, making direct use of textual cues for guidance challenging. \textbf{(c):} Our core insight is to leverage the strong text-image alignment capabilities of foundation grounding models by first converting textual cues into visual cues that the tracker can easily interpret, and then using them to guide the tracker.}
\label{fig_1}
\end{figure}

Focusing similarly on the alignment between text and image regions, the visual grounding task \cite{deng2021transvg} has a long history of research. 
Recently, some foundation grounding models \cite{liu2023grounding} \cite{li2022grounded} have achieved significant success, with one key factor being the large-scale text-image training dataset.
For instance, GLIP \cite{li2022grounded} is trained on 27 million image-text pairs to establish fine-grained correspondence between textual phrases and visual objects. 
In contrast, existing VLT benchmarks typically provide only a single textual cue for an entire video sequence. 
We analyze mainstream VLT training datasets, including TNL2k \cite{TNL2K}, LaSOT \cite{LaSOT}, and OTB99-Lang \cite{OTB-Lang}, and find that they contain a total of 1.2 million video frames but only 1,000 text annotations, as shown in Fig.~\ref{fig_1} (a). 
This reveals a significant imbalance between the textual and visual modalities. 
As shown in Fig.~\ref{fig_1} (b), on one hand, the relatively abundant visual data gives VLTs strong visual understanding capabilities, allowing them to effectively utilize visual cues to guide tracking. 
On the other hand, the severe scarcity of textual data constrains VLTs' ability to align textual and visual modalities, making it difficult to leverage textual cues for guiding the tracker.

\begin{figure}[htbp]
\centering
\includegraphics[width=\linewidth]{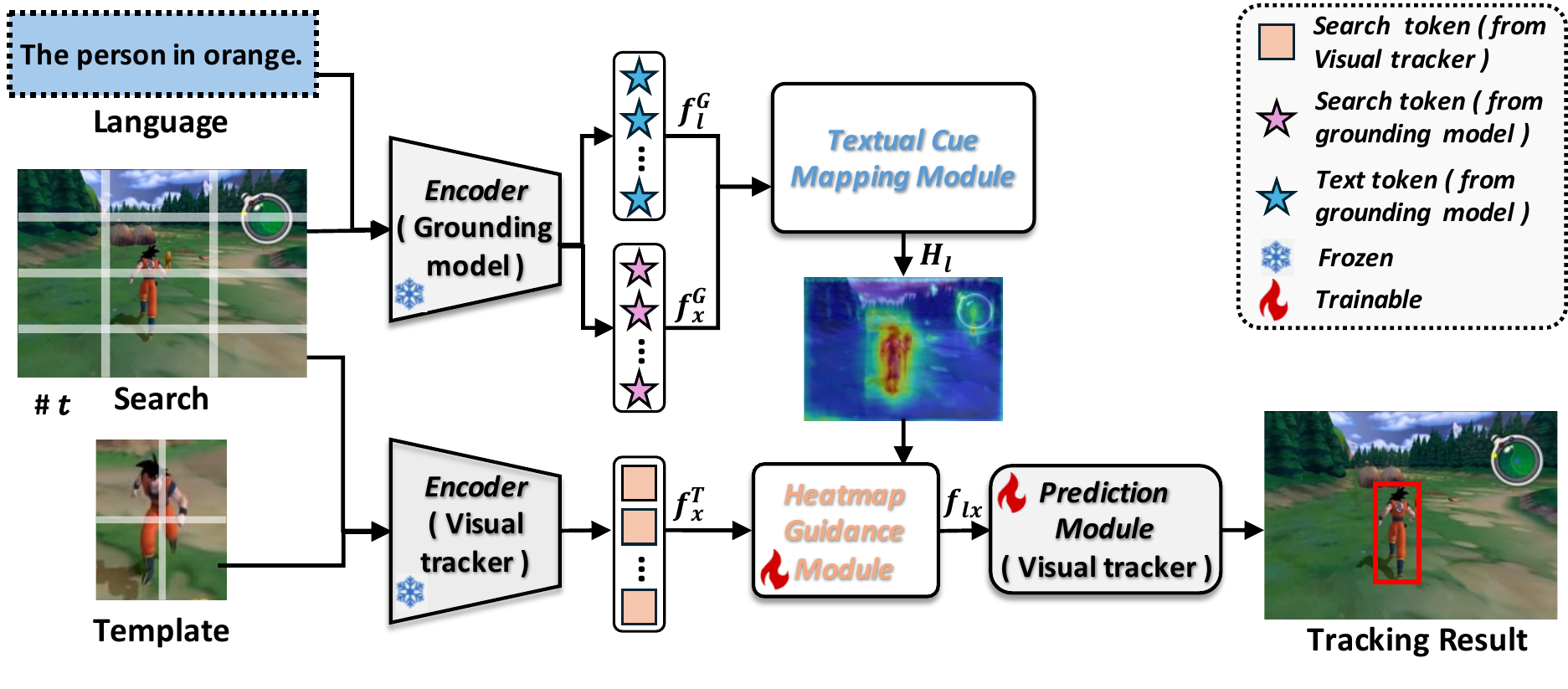}
\caption{\textbf{Framework of our vision-language tracker  (CTVLT).} Our proposed textual cue utilization method consists of the \textit{encoder module from a foundation grounding model}, along with our designed \textit{textual cue mapping module} and \textit{heatmap guidance module}. As a plug-and-play method, it can be seamlessly integrated between the \textit{encoder and prediction modules of a visual tracker}, transforming it into a vision-language tracker. }
\label{fig_2}
\end{figure}

Despite this, the ability of existing VLTs to represent and utilize textual information is developed within this constrained environment.
Specifically, SNLT \cite{SNLT} and VLT$_{\rm{TT}}$ utilize the BERT \cite{devlin2018bert} series of text encoders to extract textual features, which are then fused with image features. 
Due to the lack of alignment between textual and visual features, these models achieve only limited tracking performance.
In contrast, All-in-One \cite{zhang2023all} introduces an text-image contrastive loss to enhance the alignment between textual and visual features. 
Although it has shown some effectiveness, its training data still primarily rely on the existing tracking environment.
To mitigate the potential misleading impact of context words in textual cues, recent trackers, such as QueryNLT \cite{shao2024queryvlt}, TTCTrack \cite{mao2024textual}, and OSDT \cite{zhang2024one}, employ vision-text similarity measures to weight different words. Their reliance on the fine-grained correspondence between textual words and visual objects imposes higher alignment requirements on the VLTs.

To address these limitations, we propose a novel, plug-and-play textual cue utilization method named \textbf{CTVLT}. 
Due to the influence of existing training environments, trackers tend to have weaker capabilities in utilizing textual cues but stronger understanding of visual cues. Therefore, the core insight of \textbf{CTVLT} is to leverage the powerful text-image alignment capabilities of foundation grounding models by first \textbf{C}onverting \textbf{T}extual cues into visual cues that the tracker can easily understand, and then using these visual cues to guide the tracking process (as shown in Fig.~\ref{fig_1} (c)).

Specifically, \textbf{(i)} we design the textual cue mapping module to convert textual cues into highly interpretable visual cues. Based on the aligned text-image pairs encoded by the foundation grounding model, we propose a training-free self-attention-like refinement mechanism to generate a target distribution heatmap corresponding to the textual cues.
\textbf{(ii)} Next, we introduce the heatmap guidance module that deeply integrates the generated heatmap with the search image, ultimately enabling the effective utilization of textual cues.
\textbf{(iii)} We seamlessly integrate this innovative textual cue utilization method into the visual tracker, transforming it into a corresponding vision-language tracker. Extensive experiments conducted on mainstream benchmarks (\ie, MGIT \protect\cite{hu2023multi}, TNL2K \protect\cite{TNL2K}, and LaSOT \protect\cite{LaSOT}) demonstrate that our method significantly outperforms existing state-of-the-arts (SOTA).

\section{Method}
Fig.~\ref{fig_2} illustrates the framework of our CTVLT, which is constructed by seamlessly integrating the proposed textual cue utilization method into a visual tracker.
First, the encoder module \(E^{T}\) of the visual tracker extracts features from both the search image and the template patch, and embeds the visual template cues into the encoded search features \(f^{T}_x\).
Meanwhile, the search image and the language description are fed into the encoder module \(E^{G}\) of the foundation grounding model, generating aligned text-image feature pairs, \ie, \(f^{G}_x\) and \(f^{G}_l\). 
Based on these, our proposed textual cue mapping module converts the textual cues into a target distribution heatmap \(H_l\). Then, the heatmap guidance module deeply fuses \(H_l\) with \(f^{T}_x\), enabling the tracker to be effectively guided by the textual cues. Finally, the fused search features \(f_{lx}\) are processed by the visual tracker's prediction module \(P^{T}\) to obtain the final tracking result.

In the following sections, we first provide a brief overview of the off-the-shelf visual tracker and grounding model. Then, we delve into a thorough exploration of our proposed textual cue mapping module and heatmap guidance module.

\subsection{Preliminaries}

\subsubsection{\textbf{Visual Tracker}} 
Generally, a visual tracker consists of an encoder module \(E^{T}\) and a prediction module \(P^T\) \cite{OSTrack} \cite{xie2024autoregressive} \cite{zheng2024odtrack} \cite{zhangbeyond}.
\(E^{T}\) extracts and interacts features from the search image and template patch, thereby integrating the visual template cues into the search image. 
Then, the encoded search features \(f^T_x\) are fed into \(P^T\) to predict the bbox. 
Considering that \(E^{T}\) of most modern trackers are transformer-based \cite{OSTrack} \cite{Transformer}, \(f^T_x\) typically exists as a token sequence, \ie, \(f^T_x \in \mathbb{R}^{L_{tx} \times D}\), where \(L_{tx}\) and \(D\) represent the token length and dimensionality, respectively.
Our proposed textual cue utilization method can be seamlessly integrated between \(E^{T}\) and \(P^T\), enabling the tracker to leverage textual cues.

\subsubsection{\textbf{Foundation Grounding Model}}
For grounding models with diverse architectures, this work adopts Grounding DINO \cite{liu2023grounding}, a classic foundation model. 
In brief, it first utilizes modality-specific backbones to extract image and text features, and employs a feature enhancer facilitate their interaction, resulting in aligned text-image feature pairs.
Based on these aligned features, the language-guided query selection module and cross-modality decoder further process the features to generate the final output. For more details, please refer to \cite{liu2023grounding}.

The acquisition of aligned text-image feature pairs is a key demonstration of Grounding DINO's strong text-image alignment capability. 
Our approach primarily leverages its backbone and feature enhancer, with the cascaded module denoted as \(E^{G}\).
By feeding the language description and search image into \(E^{G}\), we obtain the aligned textual cue feature \(f^G_l \in \mathbb{R}^{L_{gl} \times D'}\) and the search feature \(f^G_x \in \mathbb{R}^{L_{gx} \times D'}\).

\begin{figure}[htbp]
\centering
\includegraphics[width=\linewidth]{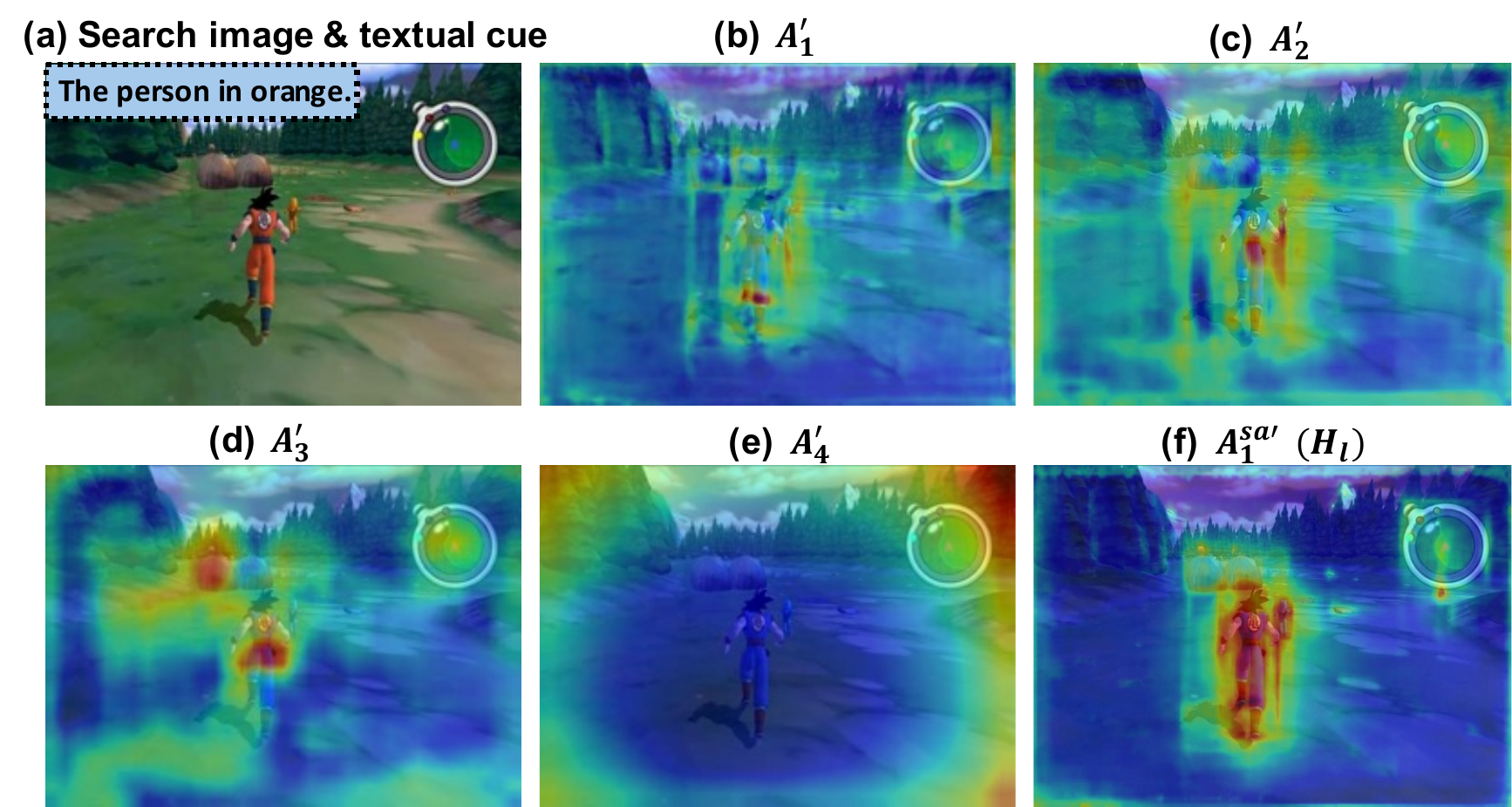}
\caption{\textbf{Visualization of different attention maps for search features with respect to textual features}.
\textbf{(a)}: Given search image and textual cue. \textbf{(b)-(e)}: Attention maps at different scales obtained through a naive process.
\textbf{(f)}: Refined result after applying our proposed method.}
\label{fig_3}
\end{figure}

\subsection{Textual Cue Mapping Module}
Although we obtain aligned feature pairs through \(E^{G}\), there remains a gap between this and our goal of effectively utilizing textual cues to guide tracking.
Based on these feature pairs, our core insight is to convert textual cues into visual cues that the tracker can easily understand, in the form of a target spatial distribution heatmap.
The process of mapping textual cues to visual cues will be detailed in the following.

To comprehensively capture image information, Grounding DINO enables interaction between textual features and multi-scale image features. 
The feature \(f^G_x\) consists of image features from four different scales, with the image tokens from each scale concatenated sequentially \cite{liu2023grounding}.
Therefore, we first need to determine which scale of image features is best suited for mapping the textual cue into the target distribution heatmap.
For that, we first calculate the attention distribution map of the image features with respect to the textual features by performing dot product operations across different scales:

\begin{equation}
  f^G_{xk} = f^G_x(S_k, E_k),
\end{equation}
\begin{equation}
  A_k = (f^G_{xk} \cdot (f^G_l)^T). \text{mean}( \text{dim} = -1).
  \label{eq_2}
\end{equation}
Here, \(f^G_{xk}\) represents the image features at the \(k\)-th scale \((1 \leq k \leq 4)\), which are extracted from \(f^G_x\) based on the start and end token indices, \(S_k\) and \(E_k\), respectively. 
Equ.~\ref{eq_2} represents the dot product operation between the image and textual features, followed by averaging along the last textual dimension.

By reshaping \(A_k \in \mathbb{R}^{L_{k} \times 1}\) into a 2D image \(A_k' \in \mathbb{R}^{w_{k} \times h_{k}}\), we can qualitatively observe the distribution of attention maps at different scales.
Through the example shown in Fig.~\ref{fig_3} (b)-(e), we can observe that the deeper the scale, the less correspondence there is between the high-attention areas and the target location.
We further conduct qualitative analysis on a large number of samples and find this to be a consistent pattern.
We hypothesize that deeper image features focus more on overall semantic information \cite{feng2023hierarchical}, while lacking in the representation of fine image details. Therefore, we choose the shallow attention map \(A_1'\) to generate subsequent heatmaps.

Although \(A_1'\) can reflect the target's spatial position, it remains relatively coarse and contains significant noise, which can mislead the tracker.
To address this, we propose a self-attention-like refinement mechanism to further filter out noise and enhance its spatial interpretability. 
Specifically, \(f^G_{x1}\) first performs correlation operations between tokens with different semantic relationships \cite{luo2024emergent} \cite{bousselham2024grounding}.
\begin{equation}
  f^{sa}_{x1} = f^G_{x1} \cdot (f^G_{x1})^T \cdot f^G_{x1}.
\end{equation}
Next, \(f^{sa}_{x1}\) and \(f^G_l\) perform the dot product operation:
\begin{equation}
  A^{sa}_1 = (f^{sa}_{x1} \cdot (f^G_l)^T). \text{mean}( \text{dim} = -1).
\end{equation}
Fig.~\ref{fig_3} (f) shows the result of reshaping \(A^{sa}_1\) into a 2D form \(A^{sa}_1\prime\). 
Clearly, compared to Fig.~\ref{fig_3} (b)-(e), the target region corresponding to the text is more distinctly highlighted, and the background noise is suppressed. Therefore, we treat \(A^{sa}_1\prime\) as our visual heatmap cue derived from the textual cue, denoted as \(H_l \in \mathbb{R}^{1 \times w_{1} \times h_{1}}\), which will be used to guide the subsequent tracking process.

\subsection{Heatmap Guidance Module}
In this section, we will explain how to leverage the target distribution heatmap \(H_l\). To achieve effective guidance, \(H_l\) needs to fully integrate with the search feature \(f^T_{x}\).

First, to facilitate interaction with \(f^T_{x}\) in the feature space, we need to perform preliminary feature extraction on the heatmap. Specifically, we design a lightweight encoding module, CNN\(_1\), based on a convolutional neural network:
\begin{equation}
  f_{h} = \text{CNN}_1(H_l),
\end{equation}
where the feature map size of \(f_{h} \in \mathbb{R}^{64 \times w_{tx} \times h_{tx}}\) matches the token sequence length of \(f^T_{x}\), \ie, \(L_{tx} = w_{tx} \times h_{tx}\).

Next, we reshape \(f^T_{x}\) from a token sequence into the form of a feature map, denoted as \(f^T_{x}\prime \in \mathbb{R}^{D \times w_{tx} \times h_{tx}}\). This allows us to concatenate \(f_{h}\) and \(f^T_{x}\prime\) along the channel dimension, resulting in \(f_{hx} \in \mathbb{R}^{(D+64) \times w_{tx} \times h_{tx}}\). 
We then use another convolutional neural network-based module, CNN\(_2\), to further process \(f_{hx}\). Through a deeper integration of \(f_{h}\) and \(f^T_{x}\prime\), we ultimately embed the textual cues into the search features, enabling the effective utilization of textual cues.
\begin{equation}
  f_{hx}\prime = \text{CNN}_2(f_{hx}).
\end{equation}
By carefully designing the structure of the convolutional neural layers, we ensure that \(f_{hx}\prime\) has the same dimensions as \(f^T_{x}\prime\), \ie, \(f_{hx}\prime \in \mathbb{R}^{D \times w_{tx} \times h_{tx}}\). 
Finally, we convert \(f_{hx}'\) back into the form of a sequence token, denoted as \(f_{lx} \in \mathbb{R}^{L_{tx} \times D}\), and feed it into the tracker's prediction module. Since \(f_{lx}\) and \(f^T_{x}\) maintain the same dimensions, no adjustments to the tracker's model structure are needed, allowing our model to be seamlessly integrated into existing visual trackers.

\begin{table}[t]
    \centering
    \caption{Comparison with SOTA methods on three popular benchmarks: MGIT \protect\cite{hu2023multi}, TNL2K \protect\cite{TNL2K}, LaSOT \protect\cite{LaSOT}. 
    The best two results are highlighted in {\color{red}red} and {\color{blue}blue}, respectively.}
    \label{tab:results_sota}
    \begin{tabular}{l|cc|cc|cc}
    \toprule
     \multicolumn{1}{c|}{\multirow{2}{*}{Method}}
      & \multicolumn{2}{c|}{MGIT} & \multicolumn{2}{c|}{TNL2K} &\multicolumn{2}{c}{LaSOT} \\ \cline{2-7}
     & AUC & P & AUC & P & AUC & P \\
      \midrule
      Feng \protect\cite{feng2019robust} & - & - & 25.0 & 27.0 & 50.0 & 56.0 \\ 
      Feng \protect\cite{feng2020real} & - & - & 25.0 & 27.0 & 35.0 & 35.0 \\ 
      GTI \protect\cite{GTI} & - & - & - & - & 47.8 & 47.6 \\
      TNL2K-II \protect\cite{TNL2K} & - & - & 42.0 & 42.0 & 51.3 & 55.4 \\
      SNLT \protect\cite{SNLT} & 3.6 & 0.4 & - & - & 54.0 & 57.4 \\ 
      Li \protect\cite{li2022cross}&  - & - & 44.0 & 45.0 & 53.0 & - \\ 
      VLT$_{\rm{TT}}$ \protect\cite{VLT} & 46.8 & 31.8 & 54.7 & 55.3 & 67.3 & 71.5 \\ 
      TransVLT \cite{zhao2023transformer} & - & - & 56.0 & - & 66.4 & 70.8 \\ 
      JointNLT \protect\cite{zhou2023joint} & {\color{blue}61.0} & {\color{blue}44.5} & 56.9 & 58.1 & 60.4 & 63.6 \\
      TransNLT \protect\cite{wang2023unified} & - & - & 57.0 & 57.0 & 60.0 & 63.0 \\
      DecoupleTNL \protect\cite{ma2023tracking} & - & - & {56.7} & {56.0} & 71.2 &{75.3} \\
      All-in-One \protect\cite{zhang2023all} & - & - & 55.3 & 57.2 & {\color{blue}{71.7}} & {\color{blue}{78.5}} \\
      MMTrack \protect\cite{zheng2023towards} & - & - & 58.6 & 59.4 & {70.0} & 75.7 \\ 
      QueryNLT \protect\cite{shao2024queryvlt} & - & - & {56.9} & {58.1} & {59.9} &{63.5} \\
      TTCTrack \protect\cite{mao2024textual} & - & - & {58.1} & {-} & {67.6} &{-} \\ 
      OSDT \protect\cite{zhang2024one} & - & - & {\color{blue}{59.3}} & {\color{blue}{61.5}} & {64.3} &{68.6} \\ 
      OneTracker \protect\cite{hong2024onetracker} & - & - & {58.0} & {59.1} & {70.5} & 76.5 \\ 
      \textbf{Ours (CTVLT)} & {\color{red}69.2} &  {\color{red} 62.9} & {\color{red}62.2} & {\color{red}79.5} & {\color{red}72.3} &{\color{red}79.7} \\ 
    \bottomrule
    \end{tabular} 
\end{table}

\section{Experiment}
\subsection{Implementation Details}
We incorporate our textual cue utilization mechanism into AQATrack \cite{xie2024autoregressive}, a recent visual tracking baseline, to develop a vision-language tracker.
The two submodules in the heatmap guidance module, CNN\(_1\) and CNN\(_2\), each consist of three convolutional neural layers. 
The textual cue mapping module contains no trainable parameters and is entirely train-free.
As shown in Fig.~\ref{fig_2}, the encoder modules from both the grounding model and visual tracker are frozen.
We use the AdamW optimizer to train the parameters over a total of 80 epochs, with 30,000 samples randomly selected in each epoch. The model is trained on a server with four RTX-3090 GPUs.
\subsection{Comparison with SOTA Methods}
As shown in Tab.~\ref{tab:results_sota}, we evaluate CTVLT on three mainstream benchmarks and compare it with existing SOTAs.

\paragraph{\textbf{MGIT}} MGIT is a novel large-scale benchmark specifically tailored for the VLT task. To handle the complex spatio-temporal causal relationships within each sequence \cite{hu2022global}, the tracker needs to leverage the textual cues as effectively as possible.
In Tab.~\ref{tab:results_sota}, CTVLT demonstrates superior performance compared to other VLTs. Particularly, it excels over the SOTA tracker JointNLT \cite{zhou2023joint}, surpassing it by 8.2\%, and 18.4\% in area under the curve (AUC) and precision (P) score, respectively. 
Unlike JointNLT, which directly utilizes textual features, our approach converts them into visual cues for use.
These results demonstrate the effectiveness of our textual cue utilization mechanism.

\paragraph{\textbf{TNL2K}} TNL2K is also crafted for the VLT task. As shown in Tab.~\ref{tab:results_sota}, our method demonstrates superior performance compared to existing VLTs.
When compared with SOTA tracker OSDT \cite{zhang2024one},  which relies on the tracker's text-image alignment capability to selectively utilize text words, our approach, by effectively leveraging the alignment capabilities of foundation models, achieves gains of 18\% in P.

\paragraph{\textbf{LaSOT}} 
 Unlike MGIT and TNL2K, LaSOT is expanded from the traditional visual tracking benchmarks \cite{LaSOT} by adding text annotations. 
 Nevertheless, CTVLT still achieves outstanding performance. In Tab.~\ref{tab:results_sota}, our method achieves significant improvements of 3.2 \% in P compared to latest OneTracker \cite{hong2024onetracker}. 
 The strong performance across multiple benchmarks reflects the generalization capability \cite{chen2024revealing} of our textual cue utilization method to diverse video environments.

\begin{table}[t]
    \caption{Ablation study on textual cue utilization methods. The best result is highlighted in {\color{red}red}.}
    \label{tab:ab_1}
    \centering
    \begin{tabular}{c|c|ccc}
        \toprule
          \# & Setting & AUC & P\(_{\text{Norm}}\) & P\\
        \midrule
         1 & \textit{direct textual cue} & 60.2 & 77.5 & 63.9 \\
         2 & \textit{naive attention map} & 61.7 & 78.9 & 65.8 \\
         3 & \textit{refined heatmap} & {\color{red} 62.2} & {\color{red}79.5} & {\color{red}66.4} \\
        \bottomrule
        \end{tabular}
\end{table}

\subsection{Ablation Studies}
The core contribution of this work is the proposal of a novel textual cue utilization mechanism, which transfers the powerful text-image alignment capability of foundation grounding models to the visual tracking task. 
To demonstrate the effectiveness of this approach, we compare it with tracker variants based on other textual cue utilization methods.

In Tab.~\ref{tab:ab_1}, \#1 represents directly using textual features to guide the tracker. Similar to MMTrack \cite{zheng2023towards}, the textual features guide the tracking process through cross-attention with the search features.
In contrast,  \#2 and \#3 adopt the paradigm where textual cues are first converted into visual cues, which are then used to guide the tracking process. 
Specifically, in \#2, the attention map (\ie, \(A_1'\)) obtained through naive text-image dot operations is treated as the visual cue, while in \#3, the visual cue is the refined heatmap (\ie, \(H_l\)) generated through our self-attention-like operation.
Compared to \#2 and \#3, we find that \#1 performs the worst, indicating that our approach of converting textual cues into visual cues is feasible.
Additionally, \#3 shows a performance improvement over \#2, primarily due to our method's ability to effectively filter out noise from the naive attention map.

\section{Conclusion}
To address the challenge of leveraging textual cues in scenarios with limited textual training data, we propose an innovative plug-and-play method for utilizing such cues. By harnessing the robust text-visual alignment capabilities of foundation grounding models, we convert textual cues into visual representations, facilitating easier interpretation by the tracker. Specifically, our textual cue mapping module transforms textual inputs into spatially interpretable target distribution heatmaps, which are then integrated into the tracking process through our heatmap guidance module. Extensive experiments on mainstream benchmarks validate the effectiveness of our approach.

\textbf{Limitations and Future Work}: 
The incorporation of foundational grounding models inevitably augments the computational load, which significantly constrains the speed of the tracker. Although we observe that employing asynchronous inference can substantially enhance tracking speed without notably compromising performance, there remains a need for future endeavors to further accelerate inference speed.

\textbf{Ethical Statement}:This is a numerical simulation study for which no ethical approval was required.

\bibliographystyle{IEEEtran} 
\bibliography{main}

\end{document}